\documentclass{article}
\usepackage{amssymb}

\usepackage{times}
\usepackage{graphicx} 
\usepackage{subfigure} 

\usepackage{natbib}

\usepackage{algorithm}
\usepackage{algorithmic}

\usepackage{hyperref}

\usepackage[accepted]{icml2015} 

\icmltitlerunning{Fast clustering for statistics}

\def\bX{\mathbf{X}}

\def\bU{\mathbf{U}}
\def\cX{\mathcal{X}}  
\def\cG{\mathcal{G}}
\def\cT{\mathcal{T}}
\def\bx{\mathbf{x}}
\def\bu{\mathbf{u}}
\def\bs{\mathbf{s}}
\def\bn{\mathbf{n}}

\begin{document} 

\twocolumn[
\icmltitle{Fast clustering for scalable statistical analysis on
structured images}

\icmlauthor{Bertrand Thirion}{bertrand.thirion@inria.fr}
\icmladdress{Parietal team, INRIA, Saclay and CEA, Neurospin France}
\icmlauthor{Andr\'es Hoyos-Idrobo}{}
\icmladdress{Parietal team, INRIA, Saclay and CEA, Neurospin France}
\icmlauthor{Jonas Kahn}{}
\icmladdress{Laboratoire Paul Painlev\'e (UMR 8524), Universit\'e de Lille 1, CNRS\, Cit\'e Scientifique--B\^at. M2, 59655 Villeneuve d’Ascq Cedex France}
\icmlauthor{ Ga\"el Varoquaux}{}
\icmladdress{Parietal team, INRIA, Saclay and CEA, Neurospin France}

\icmlkeywords{High-dimensional estimators, machine learning, brain
  imaging, clustering, random projections}

\vskip 0.3in
]

\begin{abstract} 
The use of brain images as markers for diseases or behavioral
differences is challenged by the small effects size and the ensuing
lack of power, an issue that has incited researchers to rely more
systematically on large cohorts.
Coupled with resolution increases, this leads to very large datasets.
A striking example in the case of brain imaging is that of the Human
Connectome Project: 20 Terabytes of data and growing.
The resulting data deluge poses severe challenges regarding the
tractability of some processing steps (discriminant analysis,
multivariate models) due to the memory demands posed by these data.
In this work, we revisit dimension reduction approaches, such as
random projections, with the aim of replacing costly function
evaluations by cheaper ones while decreasing the memory requirements.
Specifically, we investigate the use of alternate schemes, based on
fast clustering, that are well suited for signals exhibiting a strong
spatial structure, such as anatomical and functional brain images.
Our contribution is two-fold: \emph{i)} we propose a linear-time
clustering scheme that bypasses the percolation issues inherent in
these algorithms and thus provides compressions nearly as good as
traditional quadratic-complexity variance-minimizing clustering
schemes; \emph{ii)} we show that cluster-based compression can have
the virtuous effect of removing high-frequency noise, actually
improving subsequent estimations steps.
As a consequence, the proposed approach yields very accurate models on
several large-scale problems yet with impressive gains in
computational efficiency, making it possible to analyze large datasets.
\end{abstract}

\section{Introduction}

\paragraph{Big data in brain imaging.}
Medical images are increasingly used as markers to predict some
diagnostic or behavioral outcome.
As the corresponding biomarkers can be tenuous, researchers have come
to rely more systematically on larger cohorts to increase the power
and reliability of group studies (see e.g. \cite{Button2013} in the
case of neuroimaging).
In addition, the typical resolution of images is steadily increasing,
so that datasets become larger both in the feature and the
sample dimensions.
A striking example in the brain imaging case is that of the Human
Connectome Project (HCP): 20 Terabytes of data and growing.
The whole field is thus presently in the situation where very large
datasets are assembled. 

\paragraph{Computational bottlenecks.}
This data deluge poses severe challenges regarding the tractability of
statistical processing steps (components extraction, discriminant analysis,
multivariate models) due to the memory demands posed by the data
representations involved.
For instance, given a problem with $n$ samples and $p$ dimensions the
most classical linear algorithms (such as Principal components
analysis) have complexity $O(min(p^2n, n^2p)$, which becomes
exorbitant when both $p$ and $n$ large.
In medical imaging, $p$ would be the number of voxels (shared across
images, assuming that a prior alignment has been performed) and $n$
the number of samples:
while $p$ is e.g. of the order of $10^5-10^6$ for brain images at the
$1-2mm$ resolution, $n$ is now becoming larger ($10^6$ in the case of
the HCP dataset).
The impact on computational cost is actually worse than a simple
linear effect: as datasets no longer fit in cache, the power of
standard computational architectures can no longer be leveraged,
resulting in an extremely inefficient use of resources.
As a result, practitioners are left with the alternative of
simplifying their analysis framework or working on sub-samples of the
data (see e.g. \cite{zalesky2014}).

\paragraph{Lossy compression via random projections and clustering.}
Part of the solution to this issue is to reduce the dimensionality of
the data.
Principal components analysis, or even its randomized counterpart
\cite{halko2009}, is no longer an option, because these procedures
become inefficient due to cache size effects.
Non-linear data representations (multi-dimensional scaling, Isomap,
Laplacian eigenmaps...)  suffer from the same issue.
However, more aggressive reductions can be obtained with random projections,
i.e. the construction of random representations of the dataset in
a $k-$dimensional space, $k \ll p$. 
An essential virtue of random projections is that they come with some
guarantees on the reconstruction accuracy
(see next section).
An important drawback is that the projected data can no longer be
embedded back in the native observation space.
Moreover, random projections are a generic approach that does not take
into account any relevant information on the problem at hand: for
instance, they ignore the spatially continuous structure of the
signals in medical images.
By contrast, spatially- and contrast-aware compression schemes are
probably better suited for medical images.
We propose here to investigate adapted clustering procedures that
respect the anatomical outline of image structures.
In practice, however, standard data-based clustering (k-means,
agglomerative) yield computationally expensive estimation procedures.
Alternatively, fast clustering procedures suffer from percolation
(where a huge cluster groups most of the voxels, while many small
clusters are obtained).

\paragraph{Our contribution}
Here we propose a novel approach for fast image compression, based on
spatial clustering of the data.
This approach is designed to solve percolation issues encountered in
these settings, in order to guarantee a good enough clustering
quality.
Our contributions are:
\begin{itemize}
\item[$\bullet$] Designing a novel fast (linear-time) clustering algorithm on a
  3D lattice (image grid) that avoids percolation.
\item[$\bullet$] Showing that, used as a data-reduction strategy, it
  effectively reduces the computational cost of kernel-based estimators
  without losing accuracy.
\item[$\bullet$] Showing that, unlike random projections, this
  approach actually has a \emph{denoising} effect, that can be
  interpreted as anisotropic smoothing of the data.
\end{itemize}

\section{Theory}

\paragraph{Accuracy of random projections.}
An important characteristic of random projections is the existence of
theorems that guarantee the accuracy of the projection, in particular
the Johnson-Lindenstrauss lemma \cite{johnson1984} and its variants:

Given $0<\varepsilon<1$, a set $\cX$ of $n$ points in $\mathbb{R}^p$,
and a number $k > \frac{8 \log{(n)}}{\varepsilon^2}$, there is a linear
map $f: \mathbb{R}^p \longrightarrow \mathbb{R}^k$ such that
\begin{equation}
(1-\varepsilon)\|\bx_1-\bx_2\|^2 \leq \|f(\bx_1) - f(\bx_2)\|^2 \leq (1+\varepsilon)\|\bx_1-\bx_2\|^2
\label{eq:isometry}
\end{equation}
for all $(\bx_1, \bx_2) \in \cX \times \cX$. The map $f$ is
simply taken as the projection to a random $k$-dimensional subspace
with rescaling.
The interpretation is that, given a large enough number of random
projections of a given dataset, one can obtain a faithful
representation with explicit control on the error.
This accurate representation (in the sense of the $\ell_2$ norm) can
then be used for further analyses, such as kernel methods that
consider between-sample similarities (see e.g. \cite{rahimi2007}).
In addition, the number of necessary projections can
be lowered if the data are actually sampled from a sub-manifold of
$\mathbb{R}^p$ \cite{baraniuk2009}.
In practice, sparse random projections are used to reduce the memory
requirements and increase their efficiency \cite{li2006}.

There are two important limitations to this approach: \emph{i)} the
random mapping from $\mathbb{R}^p$ to $\mathbb{R}^k$ cannot be
inverted in general, due to its high dimensionality; this means that
the ensuing inference steps cannot be made explicit in original data
space; \emph{ii)} this approach ignores the structure of the data,
such as the spatial continuity (or dominance of low frequencies) in
medical images.

\paragraph{Signal versus noise.}
By contrast, clustering techniques have been used quite frequently in
medical imaging as a means to compress information, with empirical
success yet in the absence of formal guarantees, as in super-voxel
approaches \cite{heinrich2013}.
The explanation is that medical images are typically composed of signal
and noise, such that the high-frequency noise is reduced by
within-cluster averaging, while the low-frequency signal of interest is
preserved. 
If we denote an image, the associated signal and noise by $\bx$, $\bs$ and
$\bn$, and by $(\bu_i)_{i\in [k]}$ a set of projectors to $k$ clusters:
\[
\bx = \bs + \bn \Longrightarrow \langle\bx,\bu_i\rangle = \langle\bs,\bu_i\rangle + \langle\bn,\bu_i\rangle \; \forall i \in [k]
\]
While $\langle\bs,\bu_i\rangle$ represents a local signal average,
$\langle\bn,\bu_i\rangle$ is reduced by averaging.
$(\langle\bx,\frac{\bu_i}{\|\bu_i\|^2}\rangle)_{i \in [k]}$ form thus a
compressed representation of $\bx$.
The problem boils down to defining a suitable partition of the image
volume, or equivalently of the associated projectors
$(\bu_i)_{i\in[k]}$, where $k$ is large.  Data-unaware clustering
partitions are obviously sub-optimal, as they do not respect the
underlying structures and lead to signal loss.
Data-driven clustering can be performed through various approaches,
such as k-means or agglomerative clustering, but they tend to be
expensive: k-means has a complexity $O(npk)$; agglomerative clustering
(based on average or complete linkage heuristics or Ward's strategy
\cite{ward1963}) is also expensive ($O(np^2)$).
Single linkage clustering is fast but suffers from percolation
issues. Percolation is a major issue, because decompositions with one
giant cluster and singletons or quasi-singletons are obviously
suboptimal to represent the input signals.

\section{Fast clustering}

\paragraph{Percolation on lattices.}
Voxel clustering should take into account the 3D lattice structure of
medical images and be based on local image statistics (e.g. local
contrasts instead of cluster-level statistical summaries) in order to
obtain linear-time algorithms.
A given dataset $\bX$ is thus represented by a graph $\cG$ with 3D
lattice topology, where edges between neighboring voxels indexed by
$i$ and $j$ are associated with a distance $\|\bx_i-\bx_j\|$ that
measures the similarity between their features.
A common observation is that random graphs on lattices display
percolation as soon as the edge density reaches a critical density
($\approx$.2488 on a regular 3D lattice), meaning that a huge cluster
will group most of the voxels, leaving only small islands apart
\cite{stauffer1992}.
While single linkage clustering suffers from percolation, a simple
variant alleviates this problem:
\begin{enumerate}
\item Generate the minimum spanning tree $\mathcal{M}$ of  $\cG$
\item Delete \emph{randomly} $(k-1)$ edges from $\mathcal{M}$ while
  avoiding to create singletons (by a test on each incident node's
  degree).
\end{enumerate}
This strategy is called \emph{rand single linkage} or, more simply
\emph{rand single}, in this paper.  
Sophisticated strategies have been proposed in the framework of
computer vision (e.g. \cite{Felzenszwalb2004}), but they have not been
designed to avoid percolation and do not make it possible to control
the number $k$ of clusters.


\begin{algorithm}[tb]{}
\caption{Fast clustering by recursive nearest neighbor agglomeration}
\label{alg:fast_cluster}
\begin{algorithmic}[1]
\REQUIRE Input image $\bX$ with shape $(p, n)$, associated topological
graph $\cT$, nearest neighbor extraction function $nn$, connected
components extraction function $cc$, desired number $k$ of clusters.
\ENSURE Clustering of the voxels $l:[p] \rightarrow [k]$ \STATE $\cG =
(\delta({\cT_{ij}}) \| \bx_i - \bx_j \|), (i,j) \in [p]\times[p]$
\COMMENT{Create weighted graph}
\STATE $l=cc(nn(\cG))$ 
\COMMENT{connected components of nearest-neighbor graph}
\STATE $q = \#(l)$
\COMMENT{number of connected components}
\STATE $\bU = (\delta(j==l(i)), (i,j) \in [p]\times[q]$
\COMMENT{assignment matrix}
\WHILE{$q > k$}
    \STATE $\bX \leftarrow (\bU^T\bU)^{-1}\bU^T\bX$
    \COMMENT{reduced data matrix}
    \STATE $\cT \leftarrow \bU^T \cT \bU$
    \COMMENT{reduced topological model}
    \STATE $\cG = (\delta({\cT_{ij}}) \| \bx_i - \bx_j \|), (i,j) \in [q]\times[q]$
    \COMMENT{weighted graph}
    \STATE $\lambda = cc(nn(\cG), k)$  
    \COMMENT{cc extracts at most $k$ components}
    \STATE $\bU \leftarrow (\delta(j==\lambda(i)), (i,j) \in [q]\times[\#\lambda]$
    \COMMENT{assignment matrix}
    \STATE $q \leftarrow \#\lambda$
    \COMMENT{number of connected components}
    \STATE $l \leftarrow \lambda \circ l$
    \COMMENT{update the voxel labeling}
    \ENDWHILE
\end{algorithmic}
\end{algorithm}

\begin{figure}[t]
\begin{center}
\hspace*{-5mm}
\includegraphics[width=1.1\linewidth]{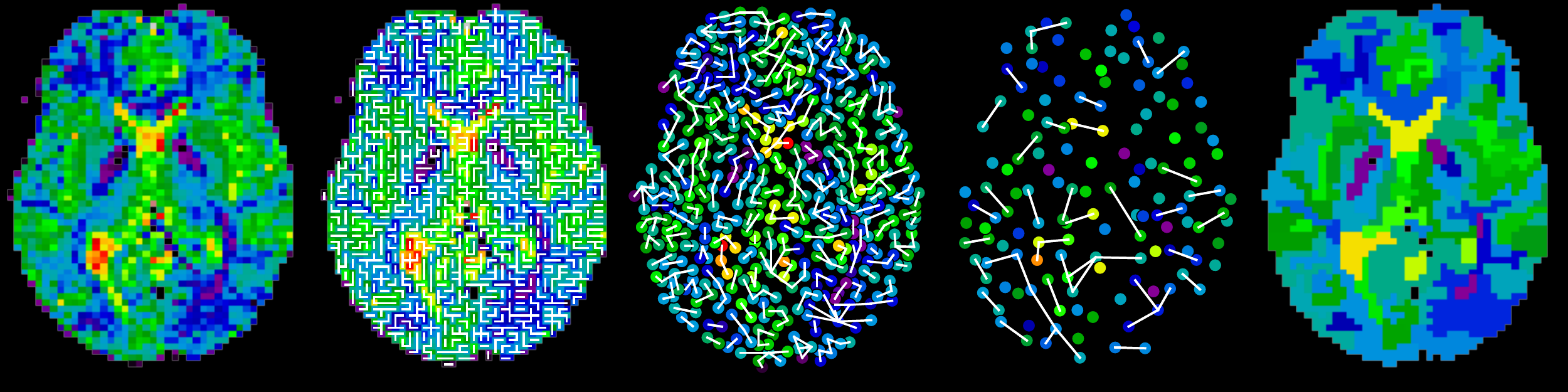}
\caption{ Principle of the fast clustering procedure illustrated
  in a real 2D brain image: the (non-percolating) nearest neighbor graph is
  computed from the origin data and so on recursively. At the last
  iteration, only the closest neighbors are associated to yield
  exactly the desired number $k$ of components.}
\label{fig:pedagogical}
\hspace*{-5mm}
\includegraphics[width=1.1\linewidth]{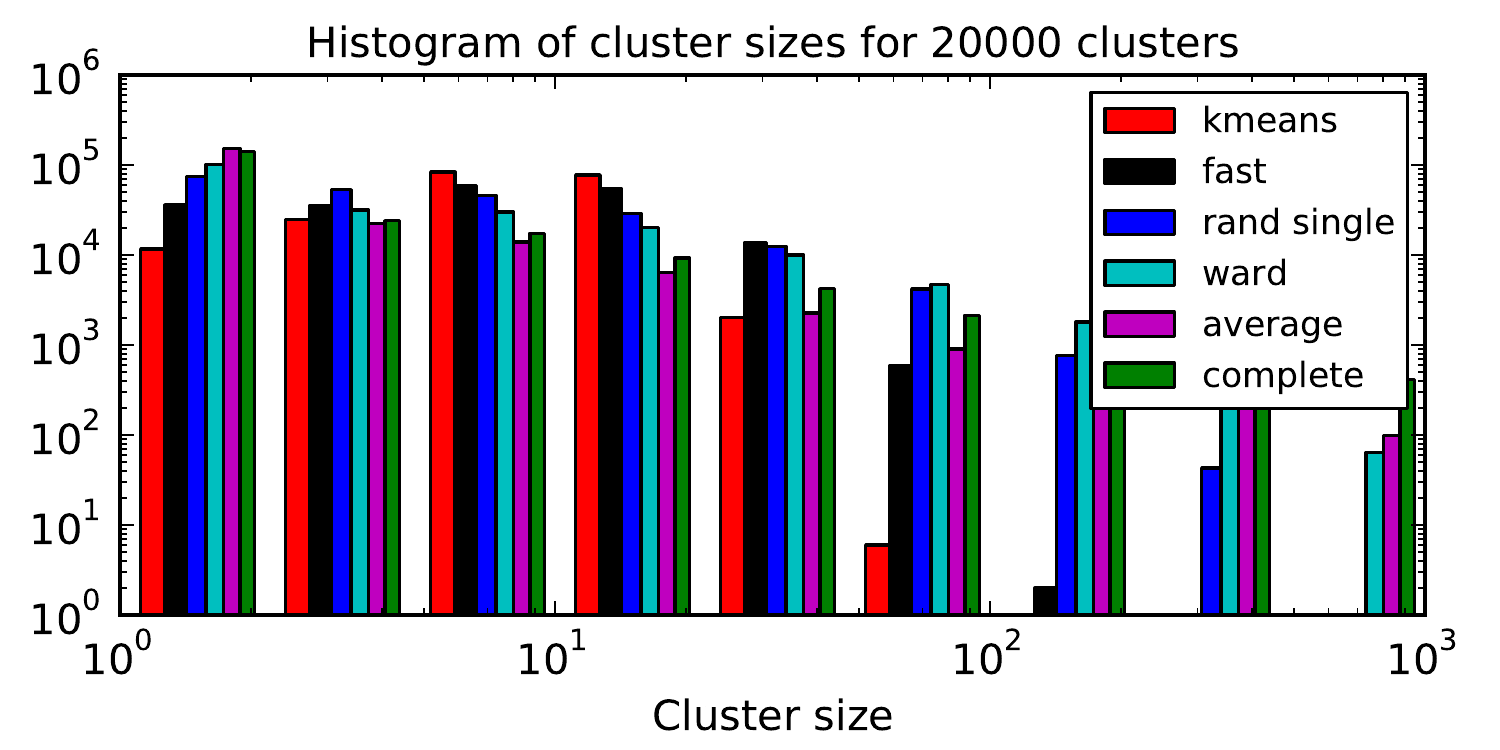}
\caption{ Percolation behavior of various clustering methods observed
  through the cluster size histogram for a fixed number $k=20,000$ of
  clusters, obtained by averaging across 10 subjects of the HCP
  dataset. \textit{K-means} and \textit{fast clustering} best avoid
  percolation, as they display neither singletons nor very large
  clusters. Traditional agglomerative clustering methods, on the other
  hand, exhibit both giant and small components. Similar results are
  obtained for other values of $k$ and datasets.}
\label{fig:percol}
\end{center}
\end{figure}

In order to obtain better clustering, we have designed the linear-time
clustering algorithm described in Alg. \ref{alg:fast_cluster} and
illustrated on a 2D brain image in Fig. \ref{fig:pedagogical}.
This algorithm is a \textit{recursive nearest-neighbor agglomeration}, that
merges clusters that are nearest neighbor of each other at each step.
Since the number of vertices is divided by at least 2 at each step,
the number of iterations is at most $O(\log(p/k))$, i.e. 5 or less in
practice, as we use typically $p/k \approx 10$ or $20$.  Since all the
operations involved are linear in the number of vertices, the
procedure is actually linear in $p$.
As predicted by theory \cite{teng2007} --namely the fact that a
one-nearest neighbor graph on any set of point (whether on a regular
lattice or not) does not exhibit percolation-- the cluster sizes are
very even.
This procedure yields more even cluster sizes than
agglomerative procedures, and performs about as well as k-means for
this purpose (see e.g. Fig. \ref{fig:percol}). We call it \textit{fast
  clustering} henceforth.

\section{Experiments}

We compare the performance of various compression schemes:
\textit{single}, \textit{average} and \textit{complete} linkage,
\textit{Ward}, \textit{fast clustering} and \textit{sparse random
  projections} in a series of tasks involving public neuroimaging
datasets (either anatomical or functional). We do not further study
k-means, as the estimation is overly expensive in the large $k$ regime
of interest.

\paragraph{Accuracy of the compressed representation}
First, we study the accuracy of the isometry in Eq. \ref{eq:isometry},
which we simply check empirically by evaluating the ratio $\eta =
\frac{\|f(\bx_1) - f(\bx_2)\|^2}{\|\bx_1 - \bx_2\|^2}$ for pairs
($\bx_1$, $\bx_2$) of samples on simulated and real data.
Random projections come with precise guarantees on the variance of
$\eta$ as a function of $k$, but no such result exists for
cluster-based representations.
The simulated data is simply a cube of shape $50 \times 50 \times 50$,
that contains a signal consisting of smooth random signal (FWHM=8mm),
with additional white noise; $n=100$ samples are drawn.
The experimental data are a sample of 10 individuals taken from NYU
test-retest resting-state functional Magnetic Resonance Imaging
(fMRI) dataset \cite{shehzad2009}, after preprocessing with a
standard SPM8 pipeline, sampled at 3mm resolution in the MNI space
($n=197$ images per subject, $p=43878$ voxels).

To avoid the bias inherent to learning the clusters and measuring the
accuracy of the representation on the same data, we perform a
cross-validation loop: the clusters a learned on a training dataset,
while the accuracy is measured on an independent dataset.
Importantly, it can be observed that clustering is actually
systematically compressive. Hence, we base our conclusions on the
variance of $\eta$ across pairs of samples, \emph{i.e.} the stability
of the ratio between distances.

\paragraph{Noise reduction}
To assess the differential effect of the spatial compressions on the
signal and the noise, we considered a set of activation
maps. 
Specifically, we relied on the motor activation dataset taken from 67
subjects of the HCP dataset \cite{barch2013}, from which we considered
the activation maps related to five different contrasts: (moving the)
\textit{left hand versus average} (activation), \textit{right hand vs.
  average}, \textit{left foot vs. average}, \textit{right foot
  vs. average} and \textit{tongue vs average}.
These activation maps have been obtained by general linear model
application upon the preprocessed data resampled at 2mm resolution in
MNI space.
From these sets of maps, in each voxel we computed the ratio of the
between-condition variance (averaged across subjects) to the
between-subject variance (averaged across conditions). 
Then we did the same on the \emph{fast cluster}-based representation.
The quotient of these two values is equal to 1 whenever the signals
are identical. Values greater than 1 indicate a denoising effect, as
the between-condition variance reflects the signal of interest while
between-subject variance is expected to reflect noise plus
between-subject variability.
We simply consider the boxplot of the log of this ratio, as a function
of the number  $k$ of components.

\paragraph{Fast logistic regression}
We performed a discriminative analysis on the OASIS dataset
\cite{marcus2007}: We used $n=403$ anatomical images and processed
them with the SPM8 software to obtain modulated grey matter density
maps sampled in the MNI space at 2mm resolution.
We used these maps to predict the gender of the subject. To achieve this,
the images were masked to an approximate average mask of the grey
matter, leaving $p=140,398$ voxels.
The voxel density values were then analyzed with an $\ell_2$-logistic
classifier, the regularization parameter of which was set by
cross-validation.
This was performed for the following methods: non-reduced data,
\emph{fast clustering}, \emph{Ward} and \emph{random projection}
reduction to either $k=4,000$ or $k=20,000$ components.
The accuracy of the procedure was assessed in a 10-fold cross
validation loop.
We measured the computation time taken to reach a stable solution by
varying the convergence control parameter.

Note that the estimation problem is rotationally invariant --i.e. the
objective function is unchanged by a rotation of the feature space--
which makes it well suited for projection-based dimension reductions.
Indeed, these can be interpreted as a kernel.

\paragraph{Fast Independent Components Analysis}
We performed an Independent Components Analysis (ICA) on resting state
fMRI from the HCP dataset, as this is a task performed routinely on
this dataset. Specifically, ICA is used to separate functional
connectivity signal from noise and obtain a spatial model of the
functional connectome \cite{smith2013}.
In the present experiment we analyzed independently data from 93
subjects. These data consist of two resting-state sessions of 1200
scans. We relied on the preprocessed data, resampled at 2mm resolution
in the MNI space. Each image represents about 1GB of data, that is
converted to a data matrix with ($p\approx 220,000$, $n=1,200$).

We performed an ICA analysis of each dataset in three settings:
\textit{i)} on the raw data, \textit{ii)} on the data compressed by
\emph{fast clustering} ($k=20,000$) and \textit{iii)} on the data
compressed by sparse random projections ($k=20,000$).
We extracted $q=40$ independent components as it is a
standard number in the literature.
Based on these analyses we investigated \textit{i)} whether the
components obtained from each dataset were similar or not before and
after clustering; \textit{ii)} How similar the components of session 1 and
session 2 were after each type of processing. This was done by matching
the components across sessions with the Hungarian algorithm, using 
the absolute value of the pairwise correlation as a
between-components similarity; \textit{iii)} the computation 
time of the ICA decomposition.

\paragraph{Implementation aspects}
The data that we used are the publicly available NYU test-retest,
OASIS and HCP datasets, for which  we used the data with the preprocessing steps
provided in the release 500-subjects release.
We relied on the Scikit-learn library \cite{pedregosa2011} (v0.15) for
machine learning tasks (ICA, logistic regression) and for Ward clustering.
We relied on the Scipy library for the agglomerative clustering
methods and the use of sparse matrices.

\section{Results}

\begin{figure*}
\begin{center}
\includegraphics[width=.8\linewidth]{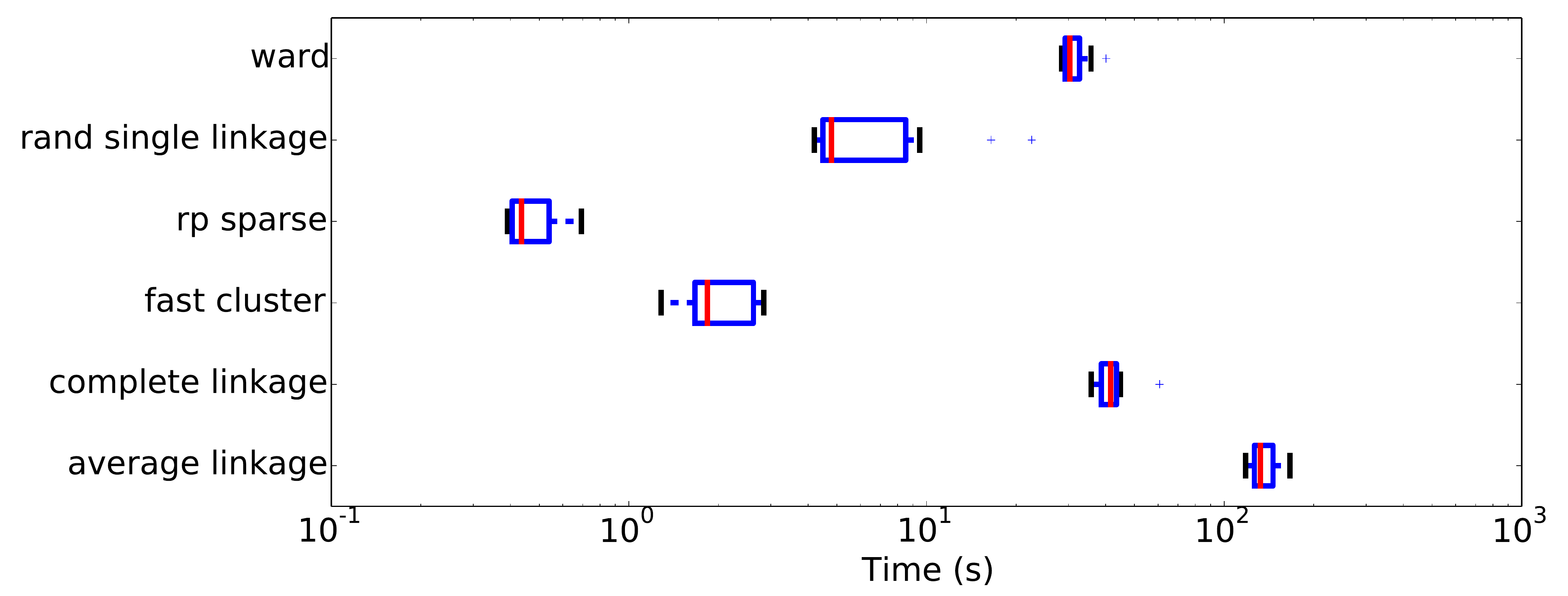}
\caption{Evaluation of the computation time of the clustering
  algorithms (to obtain $k=10,000$ clusters) tested on $n=100$ images
  taken from the OASIS dataset. The proposed fast clustering
  outperforms by far all alternatives, except random projections.}
\label{fig:time}

\begin{tabular}{cc}
\includegraphics[width=.42 \linewidth]{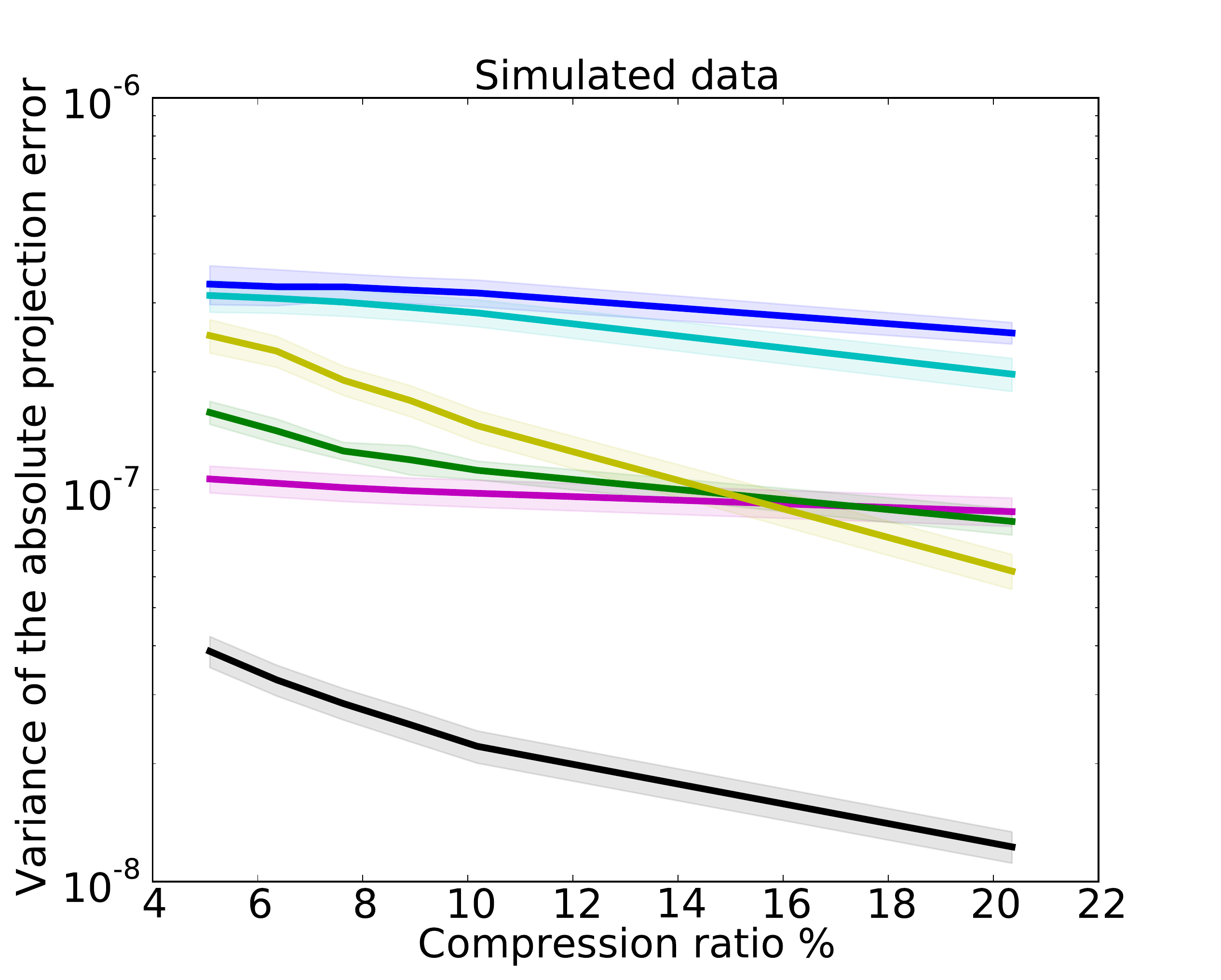} &
\includegraphics[width=.58 \linewidth]{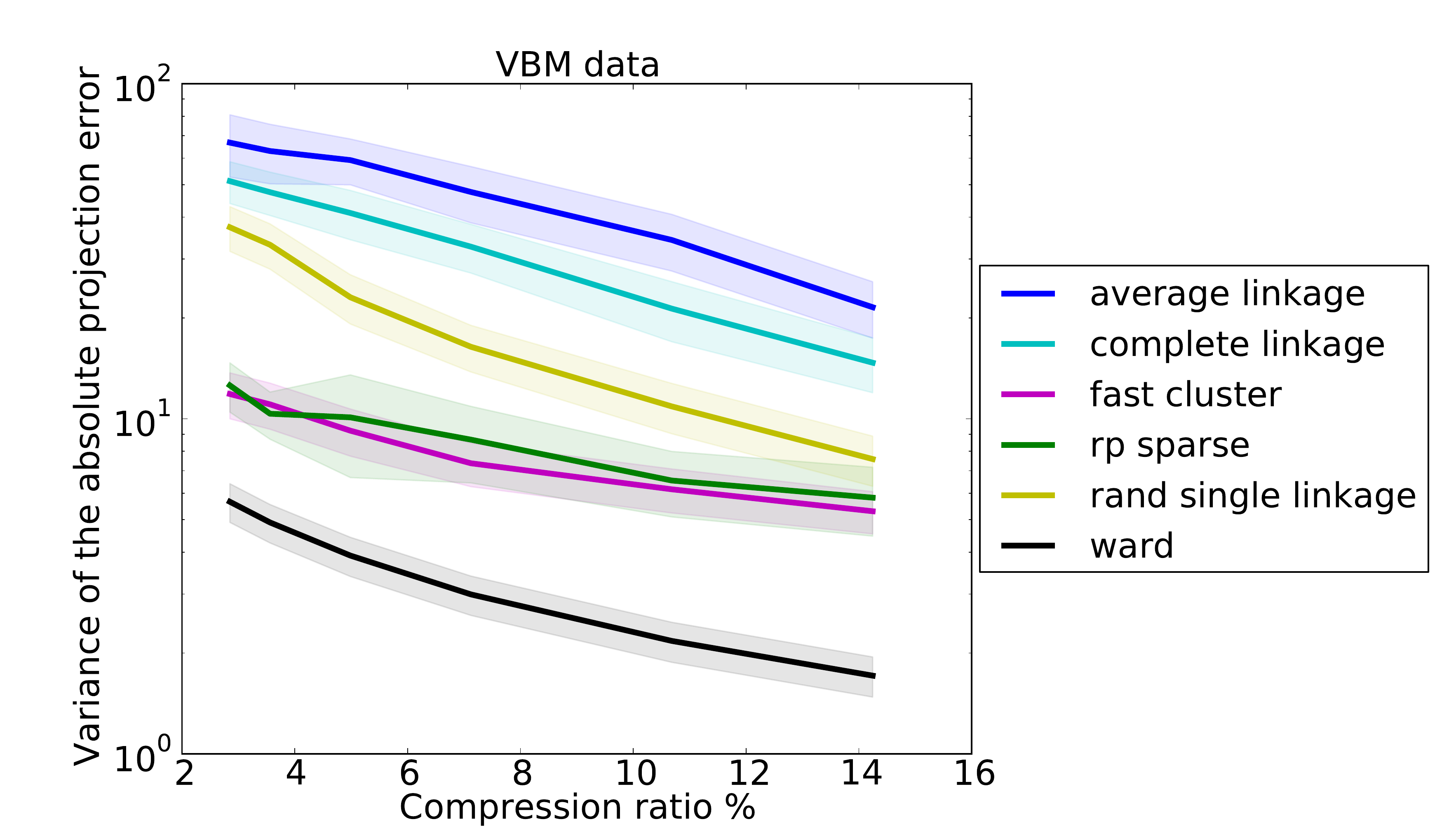}\\
\end{tabular}
\caption{Evaluation of the metric accuracy of the compressed
  representations obtained through various compression techniques, for
  different numbers of components. These experiments are based on
  simulated (left) and the OASIS dataset (right) respectively. The
  compression ratio is $\frac{k}{p}$ and error bars are across 10
  datasets;}
\label{fig:accuracy}
\vspace{5mm}
\includegraphics[width=.7\linewidth]{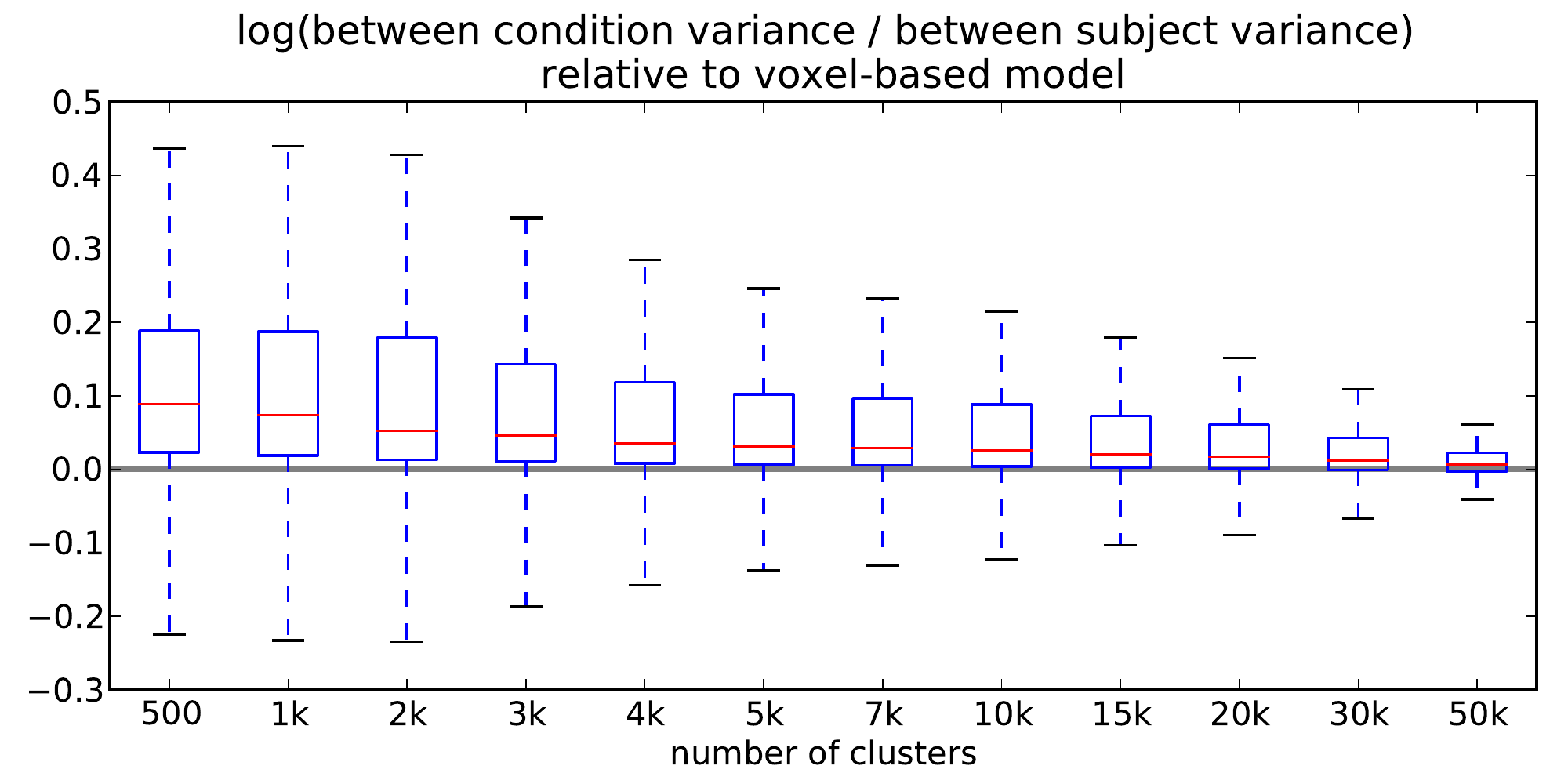}
\caption{Denoising effect of cluster-based compression: the ratio of
  between-contrasts (of interest) to between subject variance (of no
  interest) is increased when a lower number of regions is used in the
  data compression scheme. This is based on five motor
  contrasts of the HCP fMRI dataset and the \emph{fast clustering} procedure.}
\label{fig:variance}
\end{center}
\end{figure*}

\paragraph{Computational cost.}
The computational cost of the different compression schemes is
displayed in Fig. \ref{fig:time}. 
While sparse random projections are obviously faster, as they do not require
any training, \emph{fast clustering} outperforms by far Ward
clustering, which is much faster than average or complete linkage
procedures.
The clustering of a relatively large image can be obtained in a
second, this cost is actually much smaller than standard linear algebra
computations on the same dataset (blas level 3 operations).
Furthermore, this cost is reduced by learning the clustering
on a subset of the images (e.g. from 2.3s to 0.6s if one uses 10
images of OASIS instead of 100).

\paragraph{Accuracy of the compressed representation}
The quality of distance preservation is summarized in Fig.
\ref{fig:accuracy}.
The random projections accuracy improves with $k$, as predicted by
theory. 
Among the clustering algorithms, Ward clustering performs best
in terms of distance preservation. \emph{Fast clustering} performs
slightly worse, though better than random projections.
On the other hand, average and complete linkage perform poorly on this
task --which is expected, due to their tendency toward percolation.
In the next experiments, we only consider Ward and \emph{fast
  clustering}.

\paragraph{Noise reduction}
The differential effect of the spatial compression on the
signal and the noise is displayed in Fig. \ref{fig:variance}. 
This shows that, in spite of large between-voxel variability, there is
a clear trend toward a higher signal-to-noise ratio for lower values
of k.
This means that spatial compressions like clustering impose a
low-pass filtering of the data that better preserves important
discriminative features than variability components, part of which is
simply noise.

\begin{figure*}[t]
\begin{center}
\includegraphics[width=.7\linewidth]{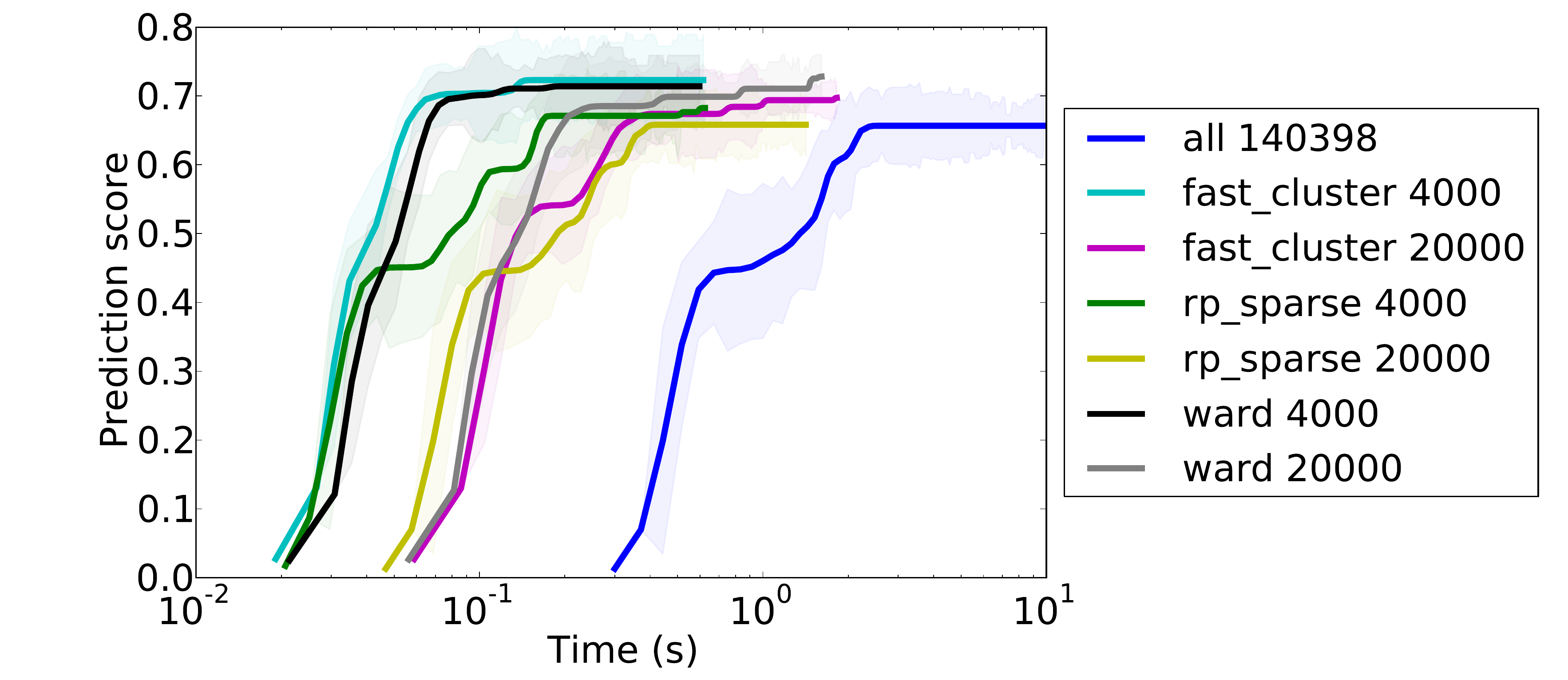}
\vspace{-3mm}
\caption{ Quality of the fit of a logistic regression of the OASIS
  dataset as a function of computation time. The cluster-based methods
  obtain significantly higher scores than regression on the whole
  dataset with a much smaller computation time (by 1.5 orders of
  magnitude). Note that the time displayed does not include cluster
  computation, which is costly in the case of Ward clustering
  ($\approx$ 10 seconds, see Fig. \ref{fig:time}).  }
\label{fig:l2log}
\includegraphics[width=.8\linewidth]{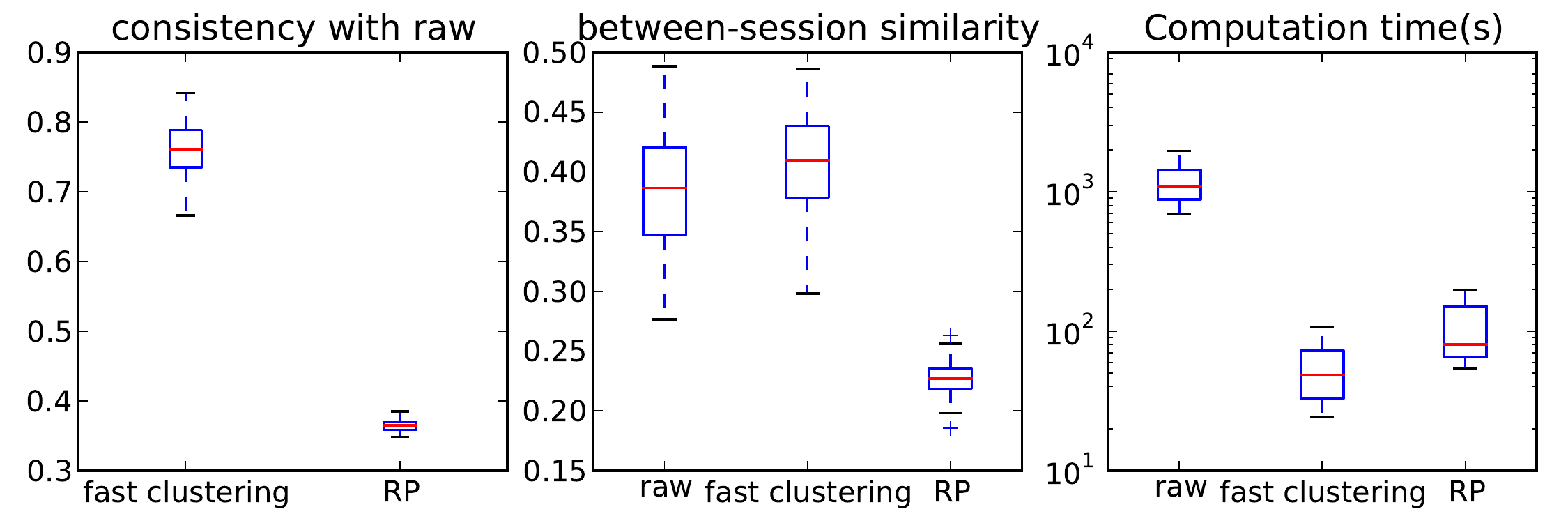}
\caption{ Results of the ICA experiments (left) the accuracy of the
  \emph{fast clustering } with respect to the non-compressed
  components is high. (Middle) across two sessions, \emph{fast
    clustering} yields components more consistent than raw data, while
  random projections fail to do so; (right) Regarding computation
  time, \emph{fast clustering} yields a gain factor of $\approx 20$,
  actually larger than $p/k$. The boxplots represent distributions
  across $93$ subjects.}
\label{fig:ica}
\end{center}
\end{figure*}

\paragraph{Fast logistic regression.}
The results of the application of logistic regression to the OASIS
dataset are displayed in Fig. \ref{fig:l2log}: this shows that the
compressed datasets (with either \emph{fast clustering}, \emph{Ward}
or \emph{random projections}) can achieve at least the same level of
accuracy as the uncompressed version, with drastic time savings.
This result is a straightforward consequence of the approximate
isometry property of the compressed representations.
The accuracy achieved is actually \emph{higher} for cluster-based
compressions than with the original data or random projections: this
illustrates again the denoising effect of spatial compression.
As a side note, achieving full convergence did not improve the
classifier performance.
Qualitatively similar results are obtained with other
rotationally invariant methods (e.g., $\ell_2$-SVMs, ridge
regression). They should carry out to any kernel machine.

\paragraph{Fast Independent Components Analysis}
The results of the ICA experiment are summarized in Fig. \ref{fig:ica}: 
We found that the $q=40$ first components were highly similar before
and after \emph{fast clustering}: the average absolute correlation
between the components was about $0.75$, while random projections do
not recover the components (average correlation $<0.4$).
Across two sessions, the components obtained by clustering are
actually \emph{more similar} after clustering than before, showing
again the denoising effect of clustering. 
This effect was observed in
all $93$ subjects, hence is extremely significant ($p<10^{-10}$,
paired Wilcoxon rank test).
On the opposite, random projections yielded a degradation of the
similarity: this is because random projections perturb the statistical
structure of the data, in particular the deviations from normality,
which are used by ICA. As a consequence, ICA cannot recover the
sources derived from the original data.
By contrast, the statistical structure is mostly preserved after
clustering.
Finally, the computation time is reduced by a factor of 20, while
$\frac{p}{k}\approx 12$ thus improving drastically the tractability of
the procedure.
Faster convergence is obtained by \textit{fast-clustering} than with
random projections.
In summary, \textit{fast clustering} not only helped to make ICA
faster, it also improved the stability of the results.
Random projections cannot be used for such a purpose.

\section{Discussion}
Our experiments have shown that on moderate size datasets, a fast
clustering technique can yield impressive gains in computation speed
for a minimal overhead to build the spatially- and contrast-aware data
compression schemes.
More importantly, the gain is found to be more than linear in various 
applications.
This comes with two other good news: even in the absence of
theoretical result, we found that spatial compression schemes perform
as well as the state-of-the-art approach in data compression for
machine learning, namely random projections.
This holds thanks to the structure of medical images, where the noise
is often observed in higher frequency components than the relevant
information.
Finally, we found that the spatial compression schemes presented here
actually have a denoising effect, yielding possibly more accurate
predictors than uncompressed version, or random projections.

Conceptually, it is tempting to compare the spatial model obtained
with \emph{fast clustering} with traditional brain parcellation
or super-voxels.
The difference lies in the interpretation: we do not view spatial
compression as a meaningful model per se, but as a way to reduce data
dimensionality without losing too much information. We will typically
set $k=p/10$ and this number is necessarily a trade-off between
computational efficiency and data fidelity.
Note that in this regime, Ward clustering is slightly more powerful in
terms of representation accuracy, but it is much slower hence
cannot be considered as a practical solution.

As shown by the ICA experiment, clustering-based compression can be
used even in tasks in which the $\ell_2$ norm preservation alone does
not guarantee a good representation.
The combination of clustering, randomization and sparsity has also
proved to be an extremely effective tool in ill-posed multivariate
estimation problems \cite{varoquaux2012icml,buhlmann2012}, hence
\emph{fast clustering} seems particularly well-suited for
these problems.

In conclusion, we have shown that a procedure using our fast
clustering method as a data reduction yields a speed up of 1.5 order
of magnitude on two real-world multivariate statistic
problems. 
Moreover, on a supervised problem, we improve the prediction
performance by using our data compression scheme, as it captures
better signal than noise.
The proposed strategy is thus extremely promising regarding the
statistical analysis of big medical image datasets, as it is perfectly
compatible with efficient online estimation methods \cite{schmidt2013}.

\paragraph{Acknowledgment.}
{\small
Data were provided in part by the Human Connectome Project, WU-Minn
Consortium (Principal Investigators: David Van Essen and Kamil
Ugurbil; 1U54MH091657) funded by the 16 NIH Institutes and Centers
that support the NIH Blueprint for Neuroscience Research; and by the
McDonnell Center for Systems Neuroscience at Washington University.
}

\bibliography{biblio}
\bibliographystyle{icml2015}

\end{document}